\documentclass{article}
\PassOptionsToPackage{numbers,sort}{natbib}
\usepackage[main, final]{neurips_2026}
\usepackage[utf8]{inputenc}
\usepackage[T1]{fontenc}
\usepackage{hyperref}
\usepackage{url}
\usepackage{booktabs}
\usepackage{amsmath,amssymb,amsfonts,bm}
\usepackage{nicefrac}
\usepackage{xcolor}
\usepackage{multirow}
\usepackage{array}
\usepackage{tabularx}
\usepackage{graphicx}
\usepackage{wrapfig}
\usepackage{float}
\usepackage{microtype}
\hypersetup{colorlinks=true,linkcolor=blue,citecolor=blue,urlcolor=blue}
\newcommand{\method}{VirtualRect}
\newcommand{\vect}[1]{\mathbf{#1}}
\newcommand{\mat}[1]{\mathbf{#1}}
\newcommand{\kstar}{\mat K^{\!*}}

\newcommand{\rscer}{RsCeR}
\newcommand{\argmin}{\mathop{\mathrm{arg\,min}}}
\title{Beyond Row Alignment: Virtual-Camera-Aware Online Stereo Rectification}

\author{%
Yizhao Peng$^{1}$ \quad Rui Gong$^{1}$ \quad Zaiwang Gu$^2$ \quad Xudong Jiang$^1$ \quad Jun Cheng$^2$\thanks{Correspondence to: cheng\_jun@a-star.edu.sg}$^\ast$\\
$^1$ Nanyang Technological University, Singapore
 \\ \quad $^2$ Institute of Advanced Intelligence and Computing, A*STAR, Singapore\\
}

\begin{document}
\maketitle

\begin{abstract}
Stereo matchers estimate depth by comparing the left and right images from a stereo camera pair, but they assume that the two cameras are accurately calibrated and rectified. In real deployments, small physical shifts after calibration, caused by camera motion, mechanical drift, vibration, or installation changes, can break this assumption and degrade depth estimation. Online stereo rectification is therefore needed to correct the image pair during operation and recover a stereo geometry where standard stereo matching can work reliably. Recent online rectification methods mainly optimize row alignment, making the left and right views of the same scene point fall on the same image row. We show that this target, while necessary, is insufficient for deployable stereo depth estimation. A rectifier may make a small set of matched points look well row-aligned, but still distort the full images in ways that hurt stereo matching: useful image regions may be cut off, large blank areas may appear, objects may be stretched or squeezed, or image scale may change unevenly across the view. We propose VirtualRect, an online stereo rectification method designed for reliable real-world deployment. Rather than treating rectification as simply making the left and right image rows line up, this method estimates a rectified virtual stereo camera and combines image-wide supervision of image warp behavior. This makes the rectified images and the camera parameters come from the same geometry, so the images used for matching and the model used for depth computation remain consistent. Experiments on multiple datasets show that, compared with the prior state-of-the-art online rectifier, VirtualRect reduces vertical flow error by 17\% on Carla-Flowguided and 7\% on Semi-Truck Highway, and lowers depth MAE by 30–41\% on Carla-Flowguided across three downstream stereo matchers.
\end{abstract}

\section{Introduction}
Online stereo rectification addresses a practical failure mode in stereo depth. Stereo matchers assume that the left and right images are well rectified: the same 3D scene point should appear on the same image row in both images, so matching can be reduced to horizontal disparity search \citep{hartley2003multiple,hartley1999rectification,fusiello2000compact}. This assumption usually comes from offline calibration. In real deployments, however, vibration, temperature changes, mechanical stress, or long-term structural deformation can shift the cameras away from their calibrated geometry. Once this happens, standard stereo matching becomes unreliable, and depth estimation degrades in systems such as mobile robots, smart glasses, and vehicle-mounted wide-baseline stereo systems  \citep{dang2009continuous,hansen2012online,ling2016online,rehder2017online,wang2023smartglasses,kumar2024flowguided,gong2025rectification}. Online rectification is therefore needed to recover usable rectified image pairs during operation. 

Recent online rectifiers correctly put row alignment at the center of the problem. FloSR trains a real-time rectifier with vertical flow as supervision  \citep{kumar2024flowguided}. \rscer{}  replaces pose-first estimation with a rectification-constrained robust estimator and rectification-specific matcher supervision \citep{gong2025rectification}. These works show that online rectification should optimize row alignment directly rather than treating rectification as a by-product of pose estimation. A rectifier may make a small set of matched points fall on the same rows, while still cutting off useful image regions, creating large blank areas, stretching objects, or changing image scale unevenly. In such cases, the row-alignment error can be small, but the rectified images can still be poor inputs for the downstream stereo matcher. 

VirtualRect keeps the benefits of explicit matching and geometric solving, but makes the solver more aware of deployment needs. First, it supervises the predicted image warps by comparing their projections of an image-wide anchor grid with those produced by the pose-derived reference image warps, extending supervision beyond the observed matches. Second, after outlier rejection, it does not treat all remaining matches equally; instead, it reduces the influence of matches that are geometrically risky for the final rectification. Third, it deploys the result through a shared virtual camera, so that the rectified images and the depth computation remain consistent. As a result, VirtualRect estimates not just a row-aligning image transformation, but a rectified stereo camera that can be directly used by a downstream stereo-depth system.

Our contributions are:
\begin{itemize}
    \item We introduce \emph{metric-consistent virtual stereo deployment}, which treats the rectified virtual camera as part of the output and evaluates whether image formation and disparity-to-depth conversion are consistent.
    \item We introduce \emph{image-wide homography-behavior supervision}, which supervises the predicted homographies by comparing their mappings of an image-wide anchor grid with those of pose-derived reference homographies, complementing pointwise correspondence residuals.
    \item We introduce \emph{solver-aware observation calibration}, which attenuates geometrically risky post-consensus matches in the final constrained solve while keeping hypothesis generation driven by appearance confidence.
    \item Our method outperforms prior online rectification approaches under standard evaluation protocols and consistently improves performance across three different downstream stereo matchers. 
\end{itemize}

\section{Related Work}
\paragraph{Pose estimation and robust two-view geometry.}
A common route to online stereo rectification is to estimate a relative pose and then derive rectifying homographies from that pose. Classical pipelines combine local features, minimal solvers, and robust estimators \citep{lowe2004sift,hartley1997eightpoint,nister2004fivepoint,fischler1981ransac}. Practical robust variants include LO-RANSAC, USAC, MAGSAC, and MAGSAC++ \citep{chum2003loransac,raguram2013usac,barath2019magsac,barath2020magsacpp}. Learned matchers improve the correspondence stage \citep{detone2018superpoint,sarlin2020superglue,sun2021loftr,wang2024efficientloftr}. Learned pose and scoring systems add neural components to geometry estimation \citep{chen2022aspanformer,edstedt2024roma,brachmann2019ngransac,zhang2019oanet}. DirectionNet and 8-point ViT are representative learning-based pose baselines \citep{chen2021directional,rockwell2022eightpoint}. These methods are powerful for camera geometry, but the objective is usually pose or epipolar consistency rather than the rectified image consumed by stereo matching. Our residual-risk module is closest in spirit to learned correspondence weighting, but it is used at a different point: after RANSAC selects a rectification-consensus set, it supplies an observation model for the final rectification solve rather than replacing the matcher or the minimal solver.

\paragraph{Stereo rectification and virtual-camera choice.}
Rectification is not unique: many homography pairs can make epipolar lines horizontal while producing different crops, distortions, and resampling patterns \citep{hartley1999rectification,hartley2003multiple,fusiello2000compact}. Classical work already recognized that legal rectifications should avoid harmful image formation: Loop and Zhang reduce distortion, Gluckman and Nayar minimize resampling effects, and Fusiello and Irsara restrict the solution to quasi-Euclidean transforms \citep{loop1999rectifying,gluckman2001rectifying,fusiello2008quasi}. Subsequent analyses connect rectification choice to local discriminability and dense matching accuracy \citep{mallon2005projective,matousek2004dataoptimal,liu2016minimumdistortion}. Online rectification brings this issue back in a deployed setting. FloSR uses vertical optical flow to train a real-time rectifier for wide-baseline stereo \citep{kumar2024flowguided}. \rscer{} directly optimizes rectification constraints with semi-dense matches and a constrained estimator \citep{gong2025rectification}. \method{} asks the next deployment question: after rows are aligned, which rectified image formation and virtual camera should the downstream stereo system actually consume? Row alignment solves the search geometry, while virtual-camera deployment determines how disparity is interpreted as metric depth. 

\paragraph{Stereo matching and depth references.}
Modern stereo matchers such as DLNR, HITNet, and RAFT-Stereo assume a rectified stereo input and convert disparity into depth using a camera model \citep{zhao2023highfrequency,tankovich2021hitnet,lipson2021raftstereo}. Recent foundation stereo models, including Foundation Stereo, Monster, and Monster++, improve robustness and generalization for stereo depth \citep{wen2025foundationstereo,cheng2025monster,cheng2025monsterpp}. Our method complements these stereo networks by selecting the rectified image formation and virtual camera used by a fixed matcher, ensuring that disparity-to-depth conversion uses the same camera model that generated the rectified images.

\begin{figure}[!t]
\centering
\includegraphics[width=0.94\textwidth]{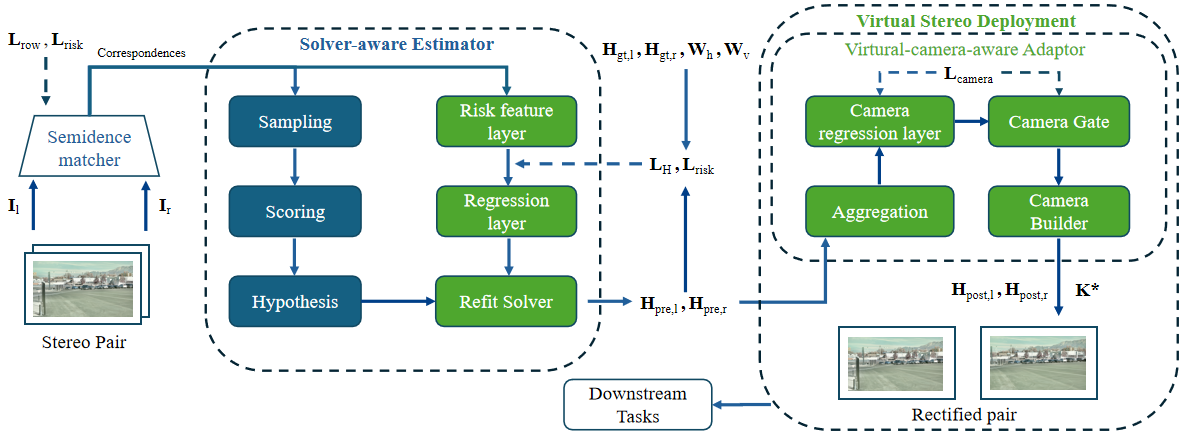}
\caption{\textbf{VirtualRect Online Rectification.} The proposed framework consists of a semi-dense matcher, a solver-aware estimator, and a virtual stereo deployment module.
The matcher takes the left and right images $\mat I_l,\mat I_r$ as input and outputs correspondences.
The solver-aware estimator samples and scores for rectification hypotheses, predicts correspondence reliability, and solves for pre-deployment rectifying homographies $\mat H^{\mathrm{pre}}_l,\mat H^{\mathrm{pre}}_r$. The virtual-camera-aware adaptor then predicts a shared rectified camera $\mat K^*$ and deploys the final homographies $\mat H^{*}_l,\mat H^{*}_r$ with $\mat K^*$ for stereo matching.}
\label{fig:overview}
\end{figure}

\section{Method}
\paragraph{Overview.}
\method{} takes a calibrated stereo pair as input and produces rectified images together with the virtual camera used for metric-depth recovery. During training, we supervise not only rectification error at matched correspondences but also how the predicted homographies map points across the image, by comparing their mappings of an image-wide anchor grid with those of pose-derived reference homographies. At inference, a semi-dense matcher first provides correspondences and appearance confidences. Next, a rectification-constrained solver estimates row-aligning homographies; RANSAC selects a consensus set, and the selected observations are calibrated using their predicted residual risks. Finally, a deployment head predicts a shared virtual camera and applies it to both views. The same virtual intrinsics are used to generate the rectified images and convert disparity to metric depth.

\subsection{Image-Wide Homography Behavior Supervision}
Pointwise row losses supervise where matches happen to be observed, but the deployed stereo matcher receives the full rectified image. A solver can fit sparse correspondences while producing a poor warp in low-texture road regions, image borders, thin structures, or other areas with weak match coverage. This mismatch is especially relevant in driving scenes, where useful depth often comes from texture-poor surfaces and object boundaries. We therefore supervise the emitted homographies as image transformations, not only the residuals of selected matches.

Let $\mathcal A$ be a regular grid of normalized anchor points covering the image and let $\pi(\cdot)$ denote homogeneous projection. The homography-behavior loss is
\begin{equation}
    \mathcal L_H=\frac{1}{|\mathcal A|}\sum_{a\in\mathcal A}
    \bigl(\|\pi(\mat H_l a)-\pi(\mat H_l^{\mathrm{gt}}a)\|_1+
    \|\pi(\mat H_r a)-\pi(\mat H_r^{\mathrm{gt}}a)\|_1\bigr).
    \label{eq:h_loss_main}
\end{equation}
The loss applies the predicted and reference homographies to the same image-wide anchor grid and measures the distance between the transformed points. Unlike the sparse row loss, which is evaluated only at matched points, the anchor grid also evaluates the warp in regions without matches.

The supervision also changes what the network is rewarded for learning. A pointwise loss can be reduced by improving matches around visually easy regions; $\mathcal L_H$ is reduced only when the final solver emits a warp whose action agrees with the reference over the image plane. This is closer to the quantity used at deployment, because the downstream matcher never observes the latent pose or the inlier set. It observes the resampled images. We therefore treat the homography pair as an image-wide operator and supervise its behavior directly.

Another way to view Eq.~\eqref{eq:h_loss_main} is as a sparse quadrature of the rectification map itself. The anchor grid does not introduce a second geometric target; it asks whether the same low-dimensional homography that fits the matches also produces a coherent image transformation over unsampled regions. This is why the term complements the constrained solver rather than weakening it: the solver still emits a physically structured rectification, while the loss makes the emitted operator visible to learning. In the ablation study, adding this term gives the largest single gain in vertical flow and pose consistency, which is why we treat it as the main geometric stabilizer of \method{}.

\subsection{Solver-Aware Observation Calibration}
For a correspondence $(\tilde{\vect u}_{l,i},\tilde{\vect u}_{r,i})$, define its vertical rectification residual as
\begin{equation}
    e_i(\mat H_l,\mat H_r)
    =
    y(\mat H_l\tilde{\vect u}_{l,i})
    -
    y(\mat H_r\tilde{\vect u}_{r,i}),
    \label{eq:rectification_residual}
\end{equation}
where $y(\cdot)$ returns the inhomogeneous vertical coordinate after homogeneous normalization. The rectifying homographies depend on the solver parameters $\theta$, so we also write this residual as $e_i(\theta)$. Robust consensus selects a plausible inlier set, but the final continuous solve still needs a reliability model for the selected observations. A uniform solver minimizes
\begin{equation}
    \hat\theta=\argmin_\theta \sum_{i\in\mathcal I} e_i(\theta)^2,
    \label{eq:uniform_refit}
\end{equation}
where $\theta$ parameterizes the five-degree-of-freedom constrained rectifier and $\mathcal I$ is the RANSAC inlier set. In wide-baseline driving, post-consensus observations are not homogeneous: repeated lane markings, occlusion boundaries, thin poles, rolling-shutter or synchronization artifacts, and support-boundary samples can pass a local consensus test while still biasing the continuous solve.

We therefore calibrate the observation precision used by the solver,
\begin{equation}
    \hat\theta=\argmin_\theta \sum_{i\in\mathcal I} w_i e_i(\theta)^2,
    \label{eq:weighted_refit}
\end{equation}
which yields the damped normal equation $(\mat J^\top\mat W\mat J+\lambda\mat I)\delta=-\mat J^\top\mat W\vect e$. Here, $\mat J=\partial\vect e/\partial\theta$ is the residual Jacobian, $\mat W=\operatorname{diag}(w_i)$, and $\lambda$ is the LM damping coefficient; the update is $\theta\leftarrow\theta+\delta$. Classical robust losses assign weights from residual magnitudes \citep{huber1964robust,tukey1977exploratory,barron2019general}. MAGSAC-style estimators marginalize over noise scales during hypothesis scoring \citep{barath2019magsac,barath2020magsacpp}. Our feature-conditioned calibration uses residual-risk weights to determine how strongly each selected observation affects the final continuous solve.

For each selected correspondence, a lightweight residual-risk head predicts a residual mean $\mu_i$, Normal-Inverse-Gamma parameters $(\nu_i,\alpha_i,\beta_i)$, and a utility logit $q_i$. The supervised residual target is

\begin{equation}
    r_i^{\mathrm{gt}}=y(\mat H_l^{\mathrm{gt}}\tilde{\vect u}_{l,i})-y(\mat H_r^{\mathrm{gt}}\tilde{\vect u}_{r,i}),
\end{equation}
with training loss
\begin{equation}
    \mathcal L_{\mathrm{risk}}=\frac{1}{P}\sum_i
    \ell_{\mathrm{NIG}}(r_i^{\mathrm{gt}};\mu_i,\nu_i,\alpha_i,\beta_i)
    +\lambda_e |r_i^{\mathrm{gt}}-\mu_i|(2\nu_i+\alpha_i).
    \label{eq:risk_loss_main}
\end{equation}
At inference, the learned solver weights are clipped to $[0.05,5]$. They do not replace RANSAC sampling or consensus selection and are applied only to the inliers selected by RANSAC during the final five-step LM refit.

\subsection{Metric-Consistent Virtual Stereo Deployment}
After the row-aligning solve and observation calibration, we obtain pre-deployment homographies $(\mat H_l^{\mathrm{pre}},\mat H_r^{\mathrm{pre}})$ in solver intrinsic $\mat K$. A deployed stereo matcher, however, consumes rectified images and converts disparity to metric depth with a camera. If image formation and depth conversion use inconsistent virtual intrinsics, disparity and metric depth are inconsistent. Legal row-aligning rectifications can also expose different finite support and disparity ranges. The deployed output should therefore be a virtual stereo camera, not only a homography pair.

Classical rectification leaves freedom in the rectified image plane and intrinsic \citep{hartley1999rectification,loop1999rectifying,gluckman2001rectifying}. We restrict this freedom to a bounded shared gauge with two stable parameters: a focal-scale change $\eta$ and a vertical principal-point shift $\Delta c_y$. They define
\begin{equation}
    \kstar=\begin{bmatrix}
    e^\eta f & 0 & c_x\\
    0 & e^\eta f & c_y+\Delta c_y\\
    0 & 0 & 1
    \end{bmatrix},
    \qquad
    \mat G=\kstar\mat K^{-1}.
    \label{eq:kstar}
\end{equation}
The final deployed homographies are
\begin{equation}
    \mat H_l^{\!*}=\mat G\mat H_l^{\mathrm{pre}},\qquad
    \mat H_r^{\!*}=\mat G\mat H_r^{\mathrm{pre}}.
    \label{eq:gauge_apply}
\end{equation}
The same shared gauge is applied to both views, preserving the row alignment solution while choosing the output camera in which the rectified images are sampled. Metric depth uses the same $\kstar$ through $Z=f^*B/d$. This consistency is the reason $\kstar$ is part of the output rather than auxiliary metadata: if the images are generated in one virtual camera but disparity is interpreted in another, the stereo system is internally inconsistent. Deployment adjusts only the shared focal scale and vertical principal point, preserving the relative row-alignment solution while selecting a bounded virtual camera.

The deployment head takes a 19-D summary of solver and residual-risk statistics, pooled image features, and matching, weight, and coverage statistics. A $19\!\to\!128\!\to\!128\!\to\!2$ MLP with ReLU after each hidden layer predicts the camera proposal, and a separate $9\!\to\!1$ sigmoid gate predicts $g$. The proposal layer is initialized to zero. The bounded outputs are
\begin{equation}
    \eta=0.08g\tanh(\tilde\eta),\qquad
    \Delta c_y=16g\tanh(\widetilde{\Delta c_y}),\qquad 0<g<1,
\end{equation}
so $|\eta|\leq0.08$ and $|\Delta c_y|\leq16$ pixels. The output is restricted to the shared camera gauge in Eq.~\eqref{eq:kstar}; it cannot change the relative row-alignment solution independently in the two views, and every change is recorded in the virtual intrinsic used for metric depth. The head is trained with three terms. The match term keeps selected correspondences row-compatible in the deployed camera,

\begin{equation}
    \mathcal L_{\mathrm{match}}=\frac{1}{|\mathcal I|}\sum_{i\in\mathcal I}
    \rho_\delta\!\left(y(\mat H_l^{\!*}\tilde{\vect u}_{l,i})-y(\mat H_r^{\!*}\tilde{\vect u}_{r,i})\right),
    \label{eq:lmatch_main}
\end{equation}
where $\rho_\delta$ is the Huber penalty. Let $S(\mat H_l,\mat H_r)$ be the common valid sampling support under the 5-pixel-border rule. The support and distortion penalties are
\begin{equation}
    \mathcal L_{\mathrm{support}}=
    \max\{0,S(\mat H_l^{\mathrm{pre}},\mat H_r^{\mathrm{pre}})-S(\mat H_l^{\!*},\mat H_r^{\!*})-\epsilon_s\},
    \quad
    \mathcal L_{\mathrm{dist}}=\eta^2+\beta(\Delta c_y/H)^2 .
    \label{eq:gauge_losses_main}
\end{equation}
The support term penalizes reductions in common valid area beyond $\epsilon_s$, while $\mathcal L_{\mathrm{dist}}$ penalizes changes in focal scale $\eta$ and vertical principal-point shift $\Delta c_y$.

\subsection{Overall Loss Functions}
Our overall loss function combines the above losses with the rectification-error loss.

\paragraph{Rectification Error Loss.}
We use the vertical rectification residual defined in Eq.~\eqref{eq:rectification_residual}. With known intrinsics, a calibrated rectifier can be written as
\begin{equation}
    \mat H_l=\mat K\mat R_l\mat K_l^{-1},\qquad
    \mat H_r=\mat K\mat R_r\mat K_r^{-1},
    \label{eq:rectifier_form}
\end{equation}
where $\mat K_l$ and $\mat K_r$ are the original intrinsics, $\mat K$ is the shared target intrinsic, and $\mat R_l,\mat R_r$ rotate the two image planes into a common rectified plane.

At inference time, the constrained estimator minimizes an aggregate of Eq.~\eqref{eq:rectification_residual} over the predicted correspondences within a tied-baseline five-degree-of-freedom rectifier family \citep{hansen2012online,gong2025rectification}. During training, we supervise the matcher using the ground-truth pose-induced rectifying homographies $(\mat H_l^{\mathrm{gt}},\mat H_r^{\mathrm{gt}})$. The rectification-error loss is defined as
\begin{equation}
\small
\mathcal{L}_{\rm err}
=
\frac{1}{P}
\sum_{i=1}^{P}
\left|
\frac{\mat H_{l[2]}^{\mathrm{gt}}\tilde{\vect u}_{l,i}}
     {\mat H_{l[3]}^{\mathrm{gt}}\tilde{\vect u}_{l,i}}
-
\frac{\mat H_{r[2]}^{\mathrm{gt}}\tilde{\vect u}_{r,i}}
     {\mat H_{r[3]}^{\mathrm{gt}}\tilde{\vect u}_{r,i}}
\right| ,
\label{eq:lerr}
\end{equation}
where $\mat H_{[2]}$ and $\mat H_{[3]}$ denote the second and third rows of $\mat H$, respectively.

The pointwise loss $\mathcal{L}_{\rm err}$ measures vertical residuals at matched points, whereas $\mathcal{L}_H$ compares the predicted and reference homographies over the anchor grid.

This loss encourages the matcher to produce correspondences that are compatible with the rectification constraint optimized by the solver. Since the rectification-error loss mainly measures vertical consistency after rectification, we additionally report pose diagnostics: rotation and translation-direction errors indicate whether the recovered rectifying geometry remains geometrically coherent before virtual-camera deployment.

\paragraph{Final Loss.}
The two training stages expose the three method components:
\begin{equation}
    \mathcal L_{\mathrm{solver}}=\mathcal L_{\rm err}+\mathcal L_{\mathrm{risk}}+\lambda_H\mathcal L_H,
    \qquad
    \mathcal L_{\mathrm{camera}}=\mathcal L_{\mathrm{match}}+\lambda_s\mathcal L_{\mathrm{support}}+\lambda_d\mathcal L_{\mathrm{dist}}.
    \label{eq:stage_objectives}
\end{equation}

At inference, the procedure remains explicit: match, robustly estimate a row-aligning solution, calibrate the final observation weights, and deploy the result as $(\mat H_l^{\!*},\mat H_r^{\!*},\kstar)$. The output is useful for downstream stereo provided that row alignment is maintained, sufficient image support is preserved, and the same virtual camera is used for image formation and metric-depth conversion. The three components therefore address different parts of the deployed rectification problem: operator-level supervision stabilizes the warp, observation calibration improves the final solve, and virtual-camera deployment makes the rectified images and metric depth belong to the same camera. Each component of the design acts on a different variable of the final deployed stereo object--the homography behavior, the observation precision, and the virtual camera.

\section{Experiments}

\subsection{Datasets, Protocols, and Metrics}
We use three benchmarks for the main evaluation. Carla-Flowguided and Semi-Truck Highway are the standard online rectification datasets introduced by FloSR and used by current SOTA \rscer{} \citep{kumar2024flowguided,gong2025rectification}. Carla is synthetic with pose annotations, while Semi-Truck is a real wide-baseline driving dataset captured under vibration and structural stress. We also report pose diagnostics on Perturbed SceneFlow, a synthetic SceneFlow-derived benchmark with known perturbations and dense correspondence supervision, motivated by the same need to reduce dependence on pretrained evaluation metrics \citep{mayer2016sceneflow,gong2025rectification}. Images from Carla and Semi-Truck are resized to $1024\times512$; Perturbed SceneFlow uses the SceneFlow evaluation resolution.

For the main benchmark results, we train a separate model on each benchmark's training split and evaluate it on the corresponding test split. The official splits do not share complete stereo pairs between training and testing, but do not require scenes or camera rigs to be disjoint.

Since Carla and Semi-Truck do not provide dense ground-truth correspondences, vertical flow remains the main dense rectification metric. And We employ a pretrained RAFT optical-flow model to estimate optical flow between stereo images, with the average magnitude of the vertical component used as the metric \citep{teed2020raft}. We also report sparse keypoint checks using SIFT and SuperGlue offsets. For the standard downstream-depth check, we follow the protocol of FloSR and \rscer{} by using DLNR-Middlebury as the stereo matcher \citep{kumar2024flowguided,gong2025rectification,zhao2023highfrequency};

%\paragraph{Coverage.}
We also compute disparity coverage. 
We distinguish geometric support from downstream coverage. Let $\Omega=\{1,\ldots,W\}\times\{1,\ldots,H\}$ denote the full rectified image grid, and let $\Omega_{lr}\subset\Omega$ be the output pixels whose inverse samples are valid in both original images under the 5-pixel-border rule. All compared rectified pairs use the same left-to-right disparity convention. For a downstream matcher $m$ with predicted disparity $d_m(p)$, we define disparity coverage as
\begin{equation}
    \mathrm{Cov}_m=\frac{1}{HW}\sum_{p\in\Omega} \mathbf{1}[p\in\Omega_{lr}]\,\mathbf{1}[d_m(p)>0].
    \label{eq:coverage_metric}
\end{equation}
It measures the fraction of the full rectified image on which the downstream matcher actually outputs a positive disparity. The datasets do not provide dense visibility annotations, so we do not apply a separate occlusion mask; all methods use the same border, disparity-positivity, reference-validity, and depth-threshold rules. Post-gauge depth is evaluated with the same virtual intrinsic that produced the post-gauge images.

\subsection{Implementation Details}
\label{sec:implementation_details}
We implement \method{} in PyTorch with Kornia geometric operators \citep{paszke2019pytorch,riba2020kornia}. For controlled comparison, the matcher, constrained solver, data protocol, and evaluation configuration are held fixed to the \rscer{} setting except for the learned observation calibration and virtual-camera deployment. The correspondence backbone is Efficient LoFTR pretrained on MegaDepth, with two matching layers, which offers a good accuracy--runtime trade-off for online rectification \citep{wang2024efficientloftr,li2018megadepth}. Models are trained for 16{,}000 steps with batch size 16 on a single NVIDIA A40 GPU. The learning rate is $10^{-4}$ on Carla-Flowguided and $10^{-5}$ on Perturbed SceneFlow and Semi-Truck Highway, with linear annealing and an 800-step warm-up. During training and testing, RANSAC runs for 16 iterations with inlier threshold $\tau=0.75$, and the Levenberg--Marquardt refit runs for 5 iterations. On Semi-Truck Highway, the filtering threshold is 0.3. Perturbed SceneFlow images are resized to $800\times416$. Optical-flow and keypoint-based evaluation follow the same configuration as prior online rectification work \citep{kumar2024flowguided,gong2025rectification}.

The residual-risk head input concatenates a 64-D local descriptor from each image, the matcher confidence, and the 2-D fine displacement, for 131 dimensions in total. The head uses a $131\!\to\!128\!\to\!128\!\to\!5$ MLP with ReLU after each hidden layer. Softplus-based output transforms enforce $\nu_i>0$, $\alpha_i>1$, and $\beta_i>0$.

\begin{figure*}[!t]
\centering
\includegraphics[width=0.98\textwidth]{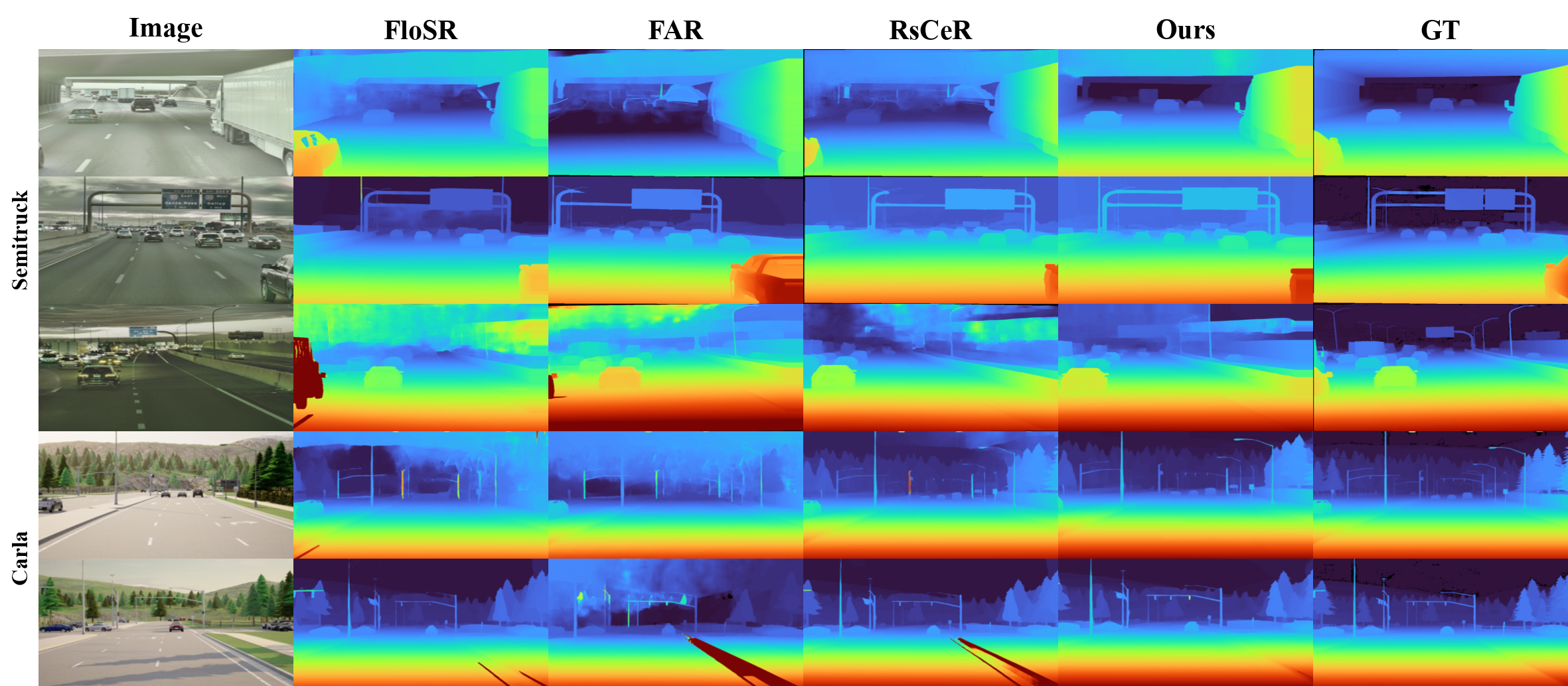}
\caption{Qualitative downstream stereo disparity comparison by different online rectification methods on Semi-Truck Highway and Carla-Flowguided. Colored panels are disparity maps; the rightmost reference uses offline-calibration rectification.}
\label{fig:qualitative_main}
\end{figure*}

Training is performed in two stages. In Stage 1, we train the residual-risk module with pointwise residual-scale supervision and image-wide homography-behavior supervision. Empirically, the latter is especially important because it constrains the learned solver by the induced warp itself, while the residual-risk term refines the observation model inside that stabilized solver. In Stage 2, we initialize the deployment head at identity and train the shared gauge correction that predicts the rectified virtual camera $\kstar$. Runtime is measured for the online rectifier at batch size 1 on one NVIDIA A40 and excludes downstream stereo inference.

\subsection{Stereo Rectification Results}
The final model is trained in two conceptual steps: residual-risk observation calibration and then virtual-camera deployment learning. Unless otherwise stated, the reported \method{} entries in the main tables are averaged over five random seeds. We organize the quantitative comparisons in the same reporting layout as \rscer{}: rectification accuracy, downstream stereo checks, and pose diagnostics. The main measurement remains row-alignment-centric: \method{} improves the vertical flow metric used by online rectification protocols, and the aligned images remain useful for downstream stereo.

\begin{table}[!t]
\centering
\small
\caption{\textbf{Stereo rectification accuracy.} We follow the keypoint-offset and vertical flow evaluation convention on Semi-Truck Highway and Carla-Flowguided. SIFT and SuperGlue offsets are independent sparse matcher checks. The \method{} row is the mean over five seeds. Arrows indicate that lower values are better.}
\label{tab:rectification_main}
\resizebox{\textwidth}{!}{
\begin{tabular}{llccc}
\toprule
Dataset & Method & SIFT Offset $\downarrow$ & SuperGlue Offset $\downarrow$ & Vertical Flow $\downarrow$ \\
 & & (pixels) & (pixels) & (pixels) \\
\midrule
\multirow{10}{*}{Semi-Truck Highway}
& SIFT + MAGSAC (2017) \citep{lowe2004sift,barath2020magsacpp} & 21.10 & 1.15 & 1.37 \\
& SP+SG + MAGSAC (2019) \citep{detone2018superpoint,sarlin2020superglue,barath2020magsacpp} & 14.97 & 1.20 & 1.51 \\
& LoFTR + MAGSAC (2021) \citep{sun2021loftr,barath2020magsacpp} & 14.94 & 0.76 & 0.86 \\
& EfficientLoFTR + MAGSAC (2024) \citep{wang2024efficientloftr,barath2020magsacpp} & 8.07 & 0.73 & 0.82 \\
& DirectionNet (2021) \citep{chen2021directional} & 8.41 & 6.10 & 6.44 \\
& ViTPose / 8-point ViT (2022) \citep{rockwell2022eightpoint} & 4.53 & 2.37 & 2.83 \\
& FAR (2024) \citep{rockwell2024far} & 4.20 & 0.49 & 0.75 \\
& FloSR w/ OF (2024) \citep{kumar2024flowguided} & 4.01 & 0.59 & 1.05 \\
& \rscer{} (2025) \citep{gong2025rectification} & 3.65 & 0.43 & 0.59 \\
& \textbf{VirtualRect (Ours)} & \textbf{3.10} & \textbf{0.42} & \textbf{0.55} \\
\midrule
\multirow{10}{*}{Carla-Flowguided}
& SIFT + MAGSAC (2017) \citep{lowe2004sift,barath2020magsacpp} & 3.60 & 0.56 & 0.59 \\
& SP+SG + MAGSAC (2019) \citep{detone2018superpoint,sarlin2020superglue,barath2020magsacpp} & 5.77 & 1.15 & 1.36 \\
& LoFTR + MAGSAC (2021) \citep{sun2021loftr,barath2020magsacpp} & 7.29 & 0.57 & 0.64 \\
& EfficientLoFTR + MAGSAC (2024) \citep{wang2024efficientloftr,barath2020magsacpp} & 3.12 & 0.59 & 0.62 \\
& DirectionNet (2021) \citep{chen2021directional} & 13.47 & 10.64 & 11.12 \\
& ViTPose / 8-point ViT (2022) \citep{rockwell2022eightpoint} & 4.34 & 1.45 & 1.72 \\
& FAR (2024) \citep{rockwell2024far} & 3.14 & 0.56 & 0.57 \\
& FloSR w/ OF (2024) \citep{kumar2024flowguided} & 3.55 & 0.64 & 1.00 \\
& \rscer{} (2025) \citep{gong2025rectification} & 3.06 & 0.42 & 0.52 \\
& \textbf{VirtualRect (Ours)} & \textbf{2.33} & \textbf{0.40} & \textbf{0.43} \\
\bottomrule
\end{tabular}}
\end{table}

\begin{table}[!t]
\centering
\scriptsize
\caption{\textbf{Downstream stereo checks and disparity coverage on Carla-Flowguided.} Each rectified input is evaluated by three stereo matchers. For each matcher, Cov is Eq.~\eqref{eq:coverage_metric}. MAE is in meters, AbsRel and Cov are percentages, and Time reports the rectifier runtime only in milliseconds per stereo pair.}
\label{tab:depth_runtime}
\setlength{\tabcolsep}{3.0pt}
\resizebox{\textwidth}{!}{
\begin{tabular}{lcccccccccc}
\toprule
\multirow{2}{*}{Method} & \multicolumn{3}{c}{Monster++} & \multicolumn{3}{c}{DLNR} & \multicolumn{3}{c}{HITNet} & \multirow{2}{*}{Time} \\
\cmidrule(lr){2-4}\cmidrule(lr){5-7}\cmidrule(lr){8-10}
& MAE $\downarrow$ & AbsRel $\downarrow$ & Cov $\uparrow$ & MAE $\downarrow$ & AbsRel $\downarrow$ & Cov $\uparrow$ & MAE $\downarrow$ & AbsRel $\downarrow$ & Cov $\uparrow$ & \\
\midrule
SIFT + MAGSAC \citep{lowe2004sift,barath2020magsacpp} & 2.03 & 5.12 & 87.05 & 2.52 & 6.49 & 88.24 & 2.67 & 9.45 & 88.21 & 367 \\
SP+SG + MAGSAC \citep{detone2018superpoint,sarlin2020superglue,barath2020magsacpp} & 8.58 & 30.22 & 71.67 & 9.89 & 31.14 & 72.69 & 10.57 & 38.04 & 72.76 & 98 \\
LoFTR + MAGSAC \citep{sun2021loftr,barath2020magsacpp} & 3.60 & 9.90 & 86.25 & 4.13 & 11.10 & 87.45 & 4.22 & 14.31 & 87.41 & 237 \\
EfficientLoFTR + MAGSAC \citep{wang2024efficientloftr,barath2020magsacpp} & 4.86 & 13.65 & 84.55 & 5.50 & 14.76 & 85.76 & 5.61 & 18.21 & 85.75 & 163 \\
FAR \citep{rockwell2024far} & 3.85 & 9.94 & 86.71 & 4.35 & 11.03 & 87.94 & 4.37 & 13.81 & 87.61 & 1672 \\
FloSR w/ OF \citep{kumar2024flowguided} & 5.72 & 20.15 & 81.39 & 6.16 & 20.23 & 82.56 & 6.43 & 25.01 & 82.52 & 81 \\
\rscer{} \citep{gong2025rectification} & 1.66 & 3.87 & 91.24 & 2.24 & 5.38 & 92.46 & 2.19 & 7.25 & 92.43 & 79 \\
\textbf{VirtualRect (Ours)} & \textbf{0.98} & \textbf{2.11} & \textbf{97.47} & \textbf{1.50} & \textbf{3.31} & \textbf{98.52} & \textbf{1.54} & \textbf{5.13} & \textbf{98.50} & 106 \\
\bottomrule
\end{tabular}}
\end{table}

Table~\ref{tab:rectification_main} reports the core rectification metrics. Relative to the previous rectification-constrained method, \method{} reduces vertical flow by 17\% on Carla-Flowguided and by 7\% on Semi-Truck Highway. The SIFT-offset improvement is an independent check because it does not reuse the dense flow model, DLNR, or the semi-dense matcher that supplies our correspondences.

Table~\ref{tab:depth_runtime} shows that the rectification improvement transfers across downstream stereo matchers rather than one evaluator on Carla-Flowguided. Relative to \rscer{}, \method{} reduces MAE by 41\%, 33\%, and 30\% for Monster++, DLNR, and HITNet, respectively. The coverage columns show the complementary effect: before metric-depth conversion, the deployed rectified images let each matcher output positive disparity over substantially more of the frame. Coverage is measured before metric-depth conversion, while the $K^*$ ablation below tests whether the image warp and the camera used for depth conversion are deployed consistently.

\begin{table}[!t]
\centering
\scriptsize
\caption{\textbf{Pose diagnostics.} These report how well the pre-gauge geometry recovers the row-alignment pose. RE is the overall rotation angular error, $R_x$, $R_y$, and $R_z$ are absolute axis-angle components, and TE is the translation-direction angular error; all are in degrees. Lower is better.}
\label{tab:pose_main}
\resizebox{\textwidth}{!}{
\begin{tabular}{lcccccccccc}
\toprule
\multirow{2}{*}{Method} & \multicolumn{5}{c}{Carla-Flowguided} & \multicolumn{5}{c}{Perturbed SceneFlow} \\
\cmidrule(lr){2-6}\cmidrule(lr){7-11}
& RE $\downarrow$ & $R_x$ $\downarrow$ & $R_y$ $\downarrow$ & $R_z$ $\downarrow$ & TE $\downarrow$ & RE $\downarrow$ & $R_x$ $\downarrow$ & $R_y$ $\downarrow$ & $R_z$ $\downarrow$ & TE $\downarrow$ \\
\midrule
EfficientLoFTR + MAGSAC \citep{wang2024efficientloftr,barath2020magsacpp} & 0.208 & 0.007 & 0.205 & 0.018 & 2.736 & 0.263 & 0.014 & 0.254 & 0.035 & 1.385 \\
FAR \citep{rockwell2024far} & 0.157 & 0.007 & 0.153 & 0.017 & 2.727 & 0.203 & 0.011 & 0.199 & 0.021 & 1.290 \\
\rscer{} \citep{gong2025rectification} & 0.081 & 0.003 & 0.077 & 0.006 & 0.860 & 0.152 & 0.008 & 0.147 & 0.018 & 0.678 \\
\textbf{VirtualRect (Ours)} & \textbf{0.021} & \textbf{0.001} & \textbf{0.016} & \textbf{0.001} & \textbf{0.258} & \textbf{0.115} & \textbf{0.005} & \textbf{0.111} & \textbf{0.014} & \textbf{0.460} \\
\bottomrule
\end{tabular}}
\end{table}

\begin{table}[!t]
\centering
\scriptsize
\caption{\textbf{Carla-trained, zero-shot on Semi-Truck.} Models are trained on Carla-Flowguided and evaluated on Semi-Truck Highway without fine-tuning. SIFT and SuperGlue report keypoint offsets, and RAFT reports vertical flow. All errors are in pixels; lower is better.}
\label{tab:carla_to_semitruck}
\begin{tabular}{lccc}
\toprule
Method & SIFT Offset $\downarrow$ & SuperGlue Offset $\downarrow$ & RAFT Vertical Flow $\downarrow$ \\
\midrule
FloSR \citep{kumar2024flowguided} & 14.00 & 5.24 & 10.65 \\
\rscer{} \citep{gong2025rectification} & 3.76 & 0.43 & 0.62 \\
\textbf{VirtualRect (Ours)} & \textbf{3.10} & \textbf{0.42} & \textbf{0.57} \\
\bottomrule
\end{tabular}
\end{table}

Table~\ref{tab:pose_main} shows that the pose diagnostics support the row-alignment half of the argument: before virtual-camera deployment, the estimated geometry is already closer to the reference geometry, reducing RE by 74\% and 24\% on Carla-Flowguided and Perturbed SceneFlow, respectively.

\paragraph{Zero-shot domain shift and baseline comparison.}
We additionally evaluate Carla-trained models on Semi-Truck Highway without fine-tuning as Table~\ref{tab:carla_to_semitruck} shows. This protocol differs from the benchmark-specific training used for Table~\ref{tab:rectification_main}. We also evaluate TESO \citep{moravec2026teso}, which introduced CARLA-Drift specifically for continuous online stereo-calibration evaluation. Existing rectification benchmarks mainly provide isolated image pairs, and no real dataset provides known time-varying decalibration together with frame-wise calibration and depth ground truth. To our knowledge, CARLA-Drift is currently the only public benchmark we found that combines long continuous stereo sequences, controlled accumulated extrinsic drift, frame-wise camera ground truth, and depth ground truth. It contains 155 sequences of 1,000 consecutive frames. The right-camera rotation follows a random walk of up to 0.01 degree per frame on each axis.

TESO maintains temporal calibration state, whereas \method{} processes each stereo pair independently, without fine-tuning or temporal state. Both methods use the same frames, undrifted references, and evaluation code. Following the benchmark definition, each \textit{-I} metric is the per-frame method error minus the corresponding undrifted-reference error; lower is better. Table~\ref{tab:carla_drift} reports VOF-I in pixels and DC-I in meters. 

\begin{table}[!t]
\centering
\scriptsize
\caption{\textbf{Zero-shot on Carla-Drift, a temporally continuous dataset.} Results on all 155 sequences (155{,}000 stereo pairs), with fixed stereo translation. Lower is better.}
\label{tab:carla_drift}
\begin{tabular}{lccc}
\toprule
Method & Temporal state & VOF-I (px) $\downarrow$ & DC-I (m) $\downarrow$ \\
\midrule
TESO \citep{moravec2026teso} & Yes & 0.07 & 0.51 \\
\textbf{VirtualRect (Ours)} & No & \textbf{0.02} & \textbf{0.25} \\
\bottomrule
\end{tabular}
\end{table}

\subsection{Ablation Study}
\label{sec:ablation}
We use two ablations to evaluate two design choices in \method{}: the estimator and the virtual-camera deployment.

Table~\ref{tab:component_ablation} isolates Stage-1 supervision while keeping the final $\kstar$ deployment fixed. The checkmark layout separates the row objective, image-wide homography-behavior supervision, and risk calibration. $\mathcal L_H$ gives the largest single improvement in both vertical flow and pose consistency, confirming that solver behavior must be constrained at the homography level. Adding $\mathcal L_{\rm risk}$ on top of this stabilized solver yields the best deployed rectification and pose metrics.

Table~\ref{tab:kstar_ablation} isolates the deployment camera. Moving from pre-gauge to post-gauge changes the rectified image formation; interpreting the post-gauge images with the old camera makes metric depth worse. Using the corresponding $\kstar$ restores consistency and improves the downstream metric. Thus online rectification should not stop at row alignment: the deployed stereo system also needs the virtual camera that generated its rectified images.

\begin{table}[!t]
\centering
\scriptsize
\caption{\textbf{Stage-1 supervision ablation on Carla-Flowguided under final deployment.} All variants use the same gauge/$\kstar$ deployment. RE and TE are angular errors in degrees. Lower is better.}
\label{tab:component_ablation}
\resizebox{0.75\textwidth}{!}{
\begin{tabular}{lccc|ccc}
\toprule
Supervision & $\mathcal L_{\rm row}$ & $\mathcal L_H$ & $\mathcal L_{\rm risk}$ & Vertical Flow $\downarrow$ & RE $\downarrow$ & TE $\downarrow$ \\
\midrule
Row alignment & $\checkmark$ & -- & -- & 0.61 & 0.036 & 0.432 \\
+ behavior supervision & $\checkmark$ & $\checkmark$ & -- & 0.48 & 0.025 & 0.288 \\
+ inlier calibration & $\checkmark$ & $\checkmark$ & $\checkmark$ & \textbf{0.43} & \textbf{0.021} & \textbf{0.258} \\
\bottomrule
\end{tabular}}
\end{table}

\begin{table}[!t]
\centering
\scriptsize
\caption{\textbf{Virtual-camera deployment ablation on Carla-Flowguided with DLNR.} Pre-gauge is the solver output; post-gauge uses the deployed virtual-camera warp. MAE is in meters and AbsRel is in percent. Lower is better.}
\label{tab:kstar_ablation}
\setlength{\tabcolsep}{3.5pt}
\resizebox{0.75\textwidth}{!}{
\begin{tabular}{lcc@{\hspace{1.2em}}lcc}
\toprule
\multicolumn{3}{c}{\rscer{}} &
\multicolumn{3}{c}{\method{}} \\
\cmidrule(lr){1-3} \cmidrule(lr){4-6}
Deployment & MAE $\downarrow$ & AbsRel (\%) $\downarrow$ &
Deployment & MAE $\downarrow$ & AbsRel (\%) $\downarrow$ \\
\midrule
pre-gauge + old $\mat K$   & 2.23 & 5.38 &
pre-gauge + old $\mat K$   & 1.61 & 3.86 \\

post-gauge + old $\mat K$  & 3.21 & 9.25 &
post-gauge + old $\mat K$  & 3.00 & 9.03 \\

post-gauge + $\kstar$      & \textbf{2.14}& \textbf{4.86}&
post-gauge + $\kstar$      & \textbf{1.50} & \textbf{3.31} \\
\bottomrule
\end{tabular}}
\end{table}
The online rectifier spends 70 ms in matching, 31 ms in RANSAC and refit, and 5 ms in the gauge and $K^*$ deployment head on one NVIDIA A40, giving 106 ms per stereo pair in total.

\section{Discussion and Limitation}
Firstly, VirtualRect assumes known input intrinsics and restricts deployment to a bounded, shared focal-scale and vertical principal-point adjustment. Current online rectification benchmarks lack dense ground-truth rectified correspondences for this deployment setting, so image-wide warp behavior is supervised using pose-derived reference homographies and downstream performance is assessed indirectly.

But to date, public real-world datasets for online stereo rectification remain scarce, although they are needed to evaluate calibration drift under real deployment conditions. While Semi-Truck provides real-world data, we do not consider its measurement precision sufficient for a definitive assessment of this problem. Additional real-world sequences with accurately measured, time-varying stereo calibration and depth ground truth are needed to establish and test real-world long-term operation or SLAM stability.

Another interesting experiment in which we select the oracle virtual-camera position separately for each stereo pair yields substantially lower measured error than the current learned predictions. This suggests that the learned virtual-camera head might still have room for improvement.

\section{Conclusion}
We presented \method{}, a deployment-consistent online stereo rectification framework that estimates a rectified virtual stereo camera rather than only a pair of row-aligning homographies. Our key observation is that reliable stereo depth requires not only row alignment, but also valid rectified image formation and a camera model that is consistent with disparity-to-depth conversion. To this end, \method{} combines image-wide homography-behavior supervision, solver-aware observation calibration, and metric-consistent virtual-camera deployment. These components stabilize the rectification solver, reduce the influence of geometrically risky matches, and ensure that the rectified images and metric depth computation share the same geometry. Experiments on standard online rectification benchmarks show that \method{} improves rectification accuracy, downstream stereo depth, and disparity coverage across multiple stereo matchers. These results suggest that online rectification should be treated as a full stereo-camera deployment problem, not merely as row alignment.

\begin{ack}
This work was supported in part by the Agency for Science, Technology and Research (A*STAR) under its MTC Programmatic Funds (Grant No. M23L7b0021) and  the National Health Innovation Centre Grant (NHIC-I2D-2502365).
\end{ack}
\clearpage

{\small
\bibliographystyle{unsrtnat}
\bibliography{main}
}
\end{document}